\documentclass[sigconf]{acmart}
\setcopyright{rightsretained}
\usepackage{booktabs}
\usepackage{amsmath}
\usepackage{array}
\usepackage{enumitem}
\usepackage{balance}
\usepackage{microtype}
\usepackage{calc}
\providecommand{\tightlist}{\setlength{\itemsep}{0pt}\setlength{\parskip}{0pt}}
\setlist[itemize]{leftmargin=*,nosep}
\setlist[enumerate]{leftmargin=*,nosep}

\acmConference[KDD-ETAAI '26]{KDD Workshop on Evaluation and Trustworthiness of Agentic AI}{August 2026}{Jeju, Republic of Korea}
\acmBooktitle{KDD Workshop on Evaluation and Trustworthiness of Agentic AI (KDD-ETAAI '26), August 2026, Jeju, Republic of Korea}
\acmYear{2026}
\acmISBN{}
\acmDOI{}

\begin{document}

\title{Reason Less, Verify More: Deterministic Gates Recover a Silent Policy-Violation Failure Mode in Tool-Using LLM Agents}

\author{Vikas Reddy}
\affiliation{%
  \institution{Independent Researcher}
  \country{}}

\author{Sumanth Reddy Challaram}
\affiliation{%
  \institution{Indian Institute of Technology Kharagpur}
  \city{Kharagpur}
  \country{India}}

\author{Abhishek Basu}
\affiliation{%
  \institution{Massachusetts Institute of Technology}
  \city{Cambridge}
  \country{USA}}

\begin{abstract}
Tool-using LLM agents can violate the very policies they are deployed to
enforce while appearing to complete the task successfully. In
policy-permissive environments, a tool may execute any well-formed call
even when the corresponding state transition is forbidden by domain
policy. The result is a silent wrong state: the booking is cancelled,
the passenger count is changed, or a user claim is acted on without
verification, and neither the tool nor the agent's self-report exposes
the violation.

We study this failure mode in the \(\tau^2\)-bench airline domain. On a
budget agent, 78\% of observed failures are silent wrong-state failures
with no tool error, and the aggregate failure rate is reproducible across disjoint
seeds rather than reflecting sampling noise. We
then evaluate a lightweight intervention: deterministic, read-only
pre-execution gates that inspect the proposed tool call and current
database state before allowing a write. A four-gate suite raises
full-benchmark success from 29.6\% to 42.0\% on gpt-4o-mini (+12.4pp;
paired task-level bootstrap \(P=0.0012\)), and the lift reproduces on a
disjoint 15-seed replication set (+12.3pp; \(P=0.0008\)). A per-gate
audit shows this lift is carried almost entirely by one high-precision
gate (100\% precision over 161 fires), while a second gate has only 5\%
precision, so gate precision must itself be audited.

The effect is concentrated where the gates actually fire: on the 26/50
firing tasks, success rises by +19.2pp, while movement on the 24
non-firing tasks does not exclude zero. Two negative controls, a
self-enforcing retail domain and BFCL, bound the mechanism: gates help
when tools are policy-permissive and add little where tools already
enforce their own preconditions. Finally, we report suggestive evidence,
not a central claim, that the same failure mode persists in a
frontier-model harness: gpt-5.2 at default reasoning still attempts
policy-violating writes, and the same gate suite improves success from
61.2\% to 71.6\% (+10.4pp; \(P=0.020\); n=5, no replication). The
contribution is a bounded evaluation and reliability result:
deterministic gates do not guarantee task success, but they can
deterministically prevent a known class of silent policy-violating
writes at the action boundary.
\end{abstract}



\maketitle
\begingroup\renewcommand\thefootnote{}\footnotetext{\footnotesize Accepted at the KDD 2026 Workshop on Evaluation and Trustworthiness of Agentic AI (KDD-ETAAI~'26). This is the authors' arXiv preprint.}\endgroup

\section{Introduction}\label{introduction}

\subsection{A trust problem, not only a capability
problem}\label{a-trust-problem-not-only-a-capability-problem}

When an LLM agent operating a customer-service tool cancels a
non-refundable reservation, modifies a passenger count that policy
forbids modifying, or processes a request on the strength of an
unverified user claim, two failures occur at once. The agent takes a
prohibited action, and because the underlying tool accepts any
syntactically valid call, the action succeeds. No exception is raised,
no error is returned, and the agent may confidently report the task
complete.

This is a trust problem. The operator observing the run sees a clean
transcript and a successful final response, but the system state is
wrong. The issue is therefore not merely whether the model can reason
about a policy in the abstract. It is whether the deployed agent system
has any reliable signal when the model fails to apply that policy before
issuing a mutating tool call.

The root cause is a property of the agent environment. In realistic tool
benchmarks such as \(\tau\)-bench and \(\tau^2\)-bench
\cite{yao2024taubench,barres2025tau2bench}, some tools are deliberately
policy-permissive: a write tool executes when its arguments are
well-formed, even if the domain policy forbids the state transition. The
policy lives in a natural-language document the model is instructed to
follow; it is not enforced by the tool. Compliance therefore depends
entirely on the model applying every relevant policy rule before every
write. When it does not, the permissive tool faithfully carries out the
violation.

On the \(\tau^2\)-bench airline tasks we study, 78\% of observed
failures are silent wrong-state failures: the final database is wrong,
but no tool ever raises an error. These are exactly the failures an
agent cannot reliably detect from its own trace.

\subsection{A reliability gap, not measurement
noise}\label{the-error-class-is-structural-rather-than-incidental}

A natural objection is that these failures are unlucky samples that more
attempts, a higher temperature, or a stronger model would wash out. For
the budget model in the airline domain, resampling is an unreliable fix,
for two reasons.

First, the failures are inconsistent rather than rare. Under the
\(\tau\)-bench unbiased pass\(_k\) estimator, success falls from
pass\(_1\) = 29.6\% to pass\(_5\) = 8.0\%: only 8\% of tasks succeed on
all five trials, so many succeed only some of the time. A single run is
therefore a poor indicator of whether the agent actually complied with
the policy.

Second, and more important, the failures are silent: no tool error marks
a violating run. Resampling can raise the chance that at least one run
succeeds, but in deployment the agent runs once, and nothing in the
trace reveals which outcome it produced, so an operator cannot retry
against a signal that never appears. This is the gap a deterministic
gate closes: rather than improving the odds across samples, it gives a
per-run guarantee that the known forbidden transition is blocked
whenever it applies. The resulting consistency gain is reported in
Section~\ref{gates-improve-reliability-under-pass_k}, and the failure
mode is not closed by scale alone
(Section~\ref{suggestive-frontier-model-evidence}).

\subsection{Deterministic pre-execution
gates}\label{deterministic-pre-execution-gates}

We study a thin layer of deterministic, read-only pre-execution gates.
Before a mutating tool call executes, a gate reads the relevant database
state and checks the proposed call against an explicit encoding of a
domain policy rule. For example, a gate may check whether a reservation
is eligible for cancellation, whether a baggage update respects cabin
and membership allowance, whether passenger count is being changed, or
whether the agent has read a record before writing to it.

A gate either allows the call or rejects it with a structured reason. It
does not write state, does not call another model, and does not use
ground-truth evaluator information. It only reads the same operational
state that the agent can read. This makes the intervention different
from output guardrails, reflection, or LLM-as-judge evaluation. The gate
adjudicates a concrete proposed state transition before mutation.

The central empirical result is that this simple mechanism recovers a
meaningful portion of silent wrong-state failures on the
\(\tau^2\)-bench airline domain, and the lift reproduces on a disjoint
replication set rather than reflecting a seed artifact
(Section~\ref{aggregate-lift-and-replication-on-the-budget-model}).

\subsection{Scope of the claim}\label{scope-of-the-claim}

The claim is intentionally bounded. We do not claim that deterministic
gates solve agent safety generally, nor that they guarantee task
success. A gate can block a violating action, and the agent can still
fail to recover. We also do not claim broad domain generality from a
single positive benchmark domain.

Instead, we make a narrower claim:

\begin{quote}
In policy-permissive tool environments with state-decidable policy rules
and final-state evaluation, silent policy-violating writes can form a
recurring failure class, and deterministic pre-execution gates can
recover a measurable fraction of those failures while providing a
deterministic guarantee over the blocked action.
\end{quote}

The rest of the paper evaluates this claim, characterizes when the lift
appears, and draws the boundary using negative controls.

\subsection{Contributions}\label{contributions}

This paper makes four contributions.

\begin{enumerate}
\def\labelenumi{\arabic{enumi}.}
\item
  \textbf{A failure-mode characterization.} We identify silent policy
  violations on policy-permissive tools as a distinct trust failure: the
  agent violates domain policy, the tool executes the forbidden write,
  and no error signal appears in the trace.
\item
  \textbf{A deterministic intervention.} We implement read-only
  pre-execution gates that check proposed mutating tool calls against
  explicit state-decidable policy predicates before execution.
\item
  \textbf{A replicated empirical result.} On the \(\tau^2\)-bench
  airline domain, a four-gate suite improves gpt-4o-mini success and
  reproduces the lift on a disjoint 15-seed replication set, with a
  per-gate audit showing the lift is carried almost entirely by a single
  high-precision gate (\texttt{cancellation\_eligibility}).
\item
  \textbf{A boundary analysis.} We show that the lift concentrates where
  gates fire, that firing is necessary but not sufficient for recovery,
  and that gates add little where tools already self-enforce their
  preconditions. We also report suggestive frontier-model evidence while
  explicitly separating it from the replicated budget-model result.
\end{enumerate}

\section{Background and Related Work}\label{background-and-related-work}

\subsection{Runtime enforcement for agentic
systems}\label{runtime-enforcement-for-agentic-systems}

A growing line of work studies runtime enforcement around LLM agents
\cite{palumbo2026formal,wang2025agentspec,yuan2026aegis,winston2026solver,qiu2025blueprint}.
Some systems compile declarative policies into reference monitors or
constraint-checking layers; others specify runtime rules as triggers and
predicates that can block unsafe actions before execution. These systems
establish that deterministic mediation at the action boundary is
feasible and that external policy checks can improve safety or
compliance.

Our work is aligned with this direction but asks a different empirical
question. Rather than treating enforcement only as a safety layer with
possible utility cost, we study when enforcement can increase benchmark
task success by preventing silent state corruption that would otherwise
make the task unrecoverable. The intervention is also deliberately
lightweight: small Python predicates over tool arguments and database
state, rather than a full provenance-graph monitor or solver-backed
policy system. Our contribution is therefore not the enforcement
mechanism, which pre-execution systems such as AEGIS \cite{yuan2026aegis}
and AgentSpec \cite{wang2025agentspec} already establish, but the
measured result on a policy-permissive, final-state benchmark:
action-boundary enforcement can raise task success, not only bound its
safety cost, and we characterize when it fires, helps, and fails.

The distinction is important. A runtime monitor may be valuable even if
it reduces task success, because it prevents unsafe actions. Here we
show a case where deterministic mediation improves both safety and
final-state task success because the blocked action was precisely the
action that would have silently corrupted the environment.

\subsection{Gates are not reflection, output rails, or
judges}\label{gates-are-not-reflection-output-rails-or-judges}

Deterministic gates are easy to conflate with three adjacent mechanisms.

First, they are not reflection. Reflection prompts ask the model to
critique or revise its own reasoning \cite{shinn2023reflexion}. Such
methods can help in some settings, but they remain model-dependent and
require either an external signal or a reliable self-correction process.
The silent wrong-state failure is difficult for reflection because the
tool produces no error.

Second, gates are not output guardrails \cite{rebedea2023nemo}. Output
rails typically screen final text or moderate generated content after
the model has responded. Our gates operate before a structured write
tool mutates state.

Third, gates are not LLM judges. An LLM judge is another stochastic
model call. A gate is a deterministic predicate over a proposed tool
call and current state. It is cheap, reproducible, and auditable.\footnote{Code and gate predicates will be released upon publication.}

\subsection{Benchmark requirements for this failure
class}\label{benchmark-requirements-for-this-failure-class}

A benchmark can evaluate this gate class only if it exposes the right
structure. We require five properties:

\begin{itemize}
\tightlist
\item
  \textbf{A1: Structured tool calls.} The proposed action must have
  machine-readable arguments.
\item
  \textbf{A2: Policy-permissive tools.} A forbidden write must execute
  silently rather than raising an error.
\item
  \textbf{A3: State-decidable policy.} The rule must be expressible as a
  deterministic predicate over current state and call arguments.
\item
  \textbf{A4: Final-state evaluation.} The evaluator must detect silent
  wrong states.
\item
  \textbf{A5: Violation-inducing tasks.} The task distribution must
  actually cause agents to violate the policy.
\end{itemize}

The \(\tau\)-bench / \(\tau^2\)-bench airline lineage satisfies all five
\cite{yao2024taubench,barres2025tau2bench}. Many other benchmarks fail
one or more axes, especially A2, A3, or A5. This scarcity is part of the
evaluation problem: many agent benchmarks measure task completion but do
not separately expose policy-permissive silent state corruption; surveys
likewise note that enterprise policy and compliance evaluation is often
underrepresented \cite{mohammadi2025survey}.

\section{The Error Class and Gate
Mechanism}\label{the-error-class-and-gate-mechanism}

\subsection{Silent policy violations on permissive
tools}\label{silent-policy-violations-on-permissive-tools}

The targeted errors are not crashes, malformed calls, or explicit
refusals. They are valid calls that mutate state while violating a
policy the model was instructed to follow.

We call this the \textbf{silent policy-violation class}. It requires a
policy-permissive tool: a tool that enforces syntax and existence checks
but does not enforce the full domain policy. For example, an airline
\texttt{cancel\_reservation} tool may check that the reservation exists
and then set its status to cancelled. If the domain policy forbids the
cancellation because of fare class, timing, insurance status, or flown
segments, the tool may still execute. The violation is silent.

A self-enforcing tool behaves differently. If a retail cancellation tool
checks that the order is pending and that the reason is valid, then a
forbidden cancellation raises an error and mutates nothing. That is a
loud and recoverable tool error, not a silent wrong state.

This distinction defines where gates are useful. On a self-enforcing
tool, a correctly written gate duplicates a check the tool already
performs. On a policy-permissive tool, a correctly written gate can
prevent a state mutation that would otherwise corrupt the final state
without any error.

\subsection{Gate definition}\label{gate-definition}

A gate is a deterministic predicate:

\[
g(\text{tool\_name}, \text{args}, \text{db\_state}) \rightarrow \{\text{allow}, \text{reject}\}.
\]

Gates are pure functions with no LLM calls and no writes. They inspect
the proposed tool call and current environment state before execution.
If a gate rejects, the underlying tool call is short-circuited and the
agent receives a structured rejection message in place of the tool
result. The agent may then re-plan.

A gate suite is evaluated in order. The first rejecting gate wins and
returns its reason. If no gate rejects, the original tool call executes
unchanged.

Our implementation is fail-open: if a gate itself raises an exception,
the harness records the error and allows the original call. This avoids
introducing new false blocks from gate implementation failures.

\subsection{The four-gate suite}\label{the-four-gate-suite}

The headline configuration is a four-gate suite (Table~\ref{tab:gates})
selected on the budget tier and held fixed across model tiers.

\begin{table*}[t]
\small
\centering
\caption{Four deterministic gates in the headline suite.}
\label{tab:gates}
\begin{tabular}[]{@{}
  >{\raggedright\arraybackslash}p{(\textwidth - 4\tabcolsep) * \real{0.3333}}
  >{\raggedright\arraybackslash}p{(\textwidth - 4\tabcolsep) * \real{0.3333}}
  >{\raggedright\arraybackslash}p{(\textwidth - 4\tabcolsep) * \real{0.3333}}@{}}
\toprule\noalign{}
\begin{minipage}[b]{\linewidth}\raggedright
Gate
\end{minipage} & \begin{minipage}[b]{\linewidth}\raggedright
Targeted write
\end{minipage} & \begin{minipage}[b]{\linewidth}\raggedright
Policy intuition
\end{minipage} \\
\midrule\noalign{}
\bottomrule\noalign{}
\texttt{cancellation\_eligibility} & \texttt{cancel\_reservation} &
Block cancellations unless the reservation is eligible under fare,
timing, insurance, and flown-segment rules. \\
\texttt{baggage\_allowance} & \texttt{update\_reservation\_baggages} &
Block baggage updates that misprice free bags or remove bags contrary to
policy. \\
\texttt{passenger\_count} & \texttt{update\_reservation\_passengers} &
Block attempts to change the passenger count, which is immutable by
policy. \\
\texttt{must\_read\_before\_write} & Any write to a record & Block
writes to records the agent has not read in the current session. \\
\end{tabular}
\end{table*}

The four gates each encode a distinct policy rule, and they contribute
unevenly. As the per-gate audit shows
(Section~\ref{gate-precision-and-load-bearing-gates}), a single
high-precision gate, \texttt{cancellation\_eligibility}, carries most of
the task-success lift, while the others extend policy coverage but are
not load-bearing on this distribution and are candidates for per-model
tuning or automatic generation. We report the four-gate suite as the
headline rather than a post-hoc single-gate optimum, and state this
concentration openly.

\subsection{Firing-share
decomposition}\label{firing-share-decomposition}

A gate can affect a trial only when it fires. Aggregate lift therefore
decomposes into a firing share and a conditional effect where firing
occurs:

\[
\Delta_{\text{aggregate}} \approx p_{\text{fire}} \times \Delta_{\text{fire}}.
\]

This decomposition is central to the analysis. If gates are truly
responsible for the lift, improvements should concentrate on tasks where
gates fire. If non-firing tasks move similarly, the measured improvement
may be unrelated to the intervention.

On the full airline benchmark with gpt-4o-mini, the gate suite produces
132 rejections, and 83/250 trials contain at least one rejection. The
aggregate improvement is +31 successful trials. Of those, +25 occur in
the firing stratum.

\section{Experimental Setup}\label{experimental-setup}

\subsection{Harness}\label{harness}

We run the \(\tau^2\)-bench airline domain as a library rather than only
through the command-line interface. This lets vanilla and gated
conditions traverse the same code path, differing only by the presence
of gates.

Each \texttt{(task,\ trial)} run builds the orchestrator, executes the
simulation, and uses the benchmark's own final-state evaluator. A trial
is counted as successful when
\texttt{reward\_info.reward\ \textgreater{}\ 0}.

We add gates through two idempotent patches to the environment.

First, a pre-call dispatcher intercepts each proposed tool call. If a
gate rejects, the original handler is not called and the state is not
mutated. If no gate rejects, execution proceeds through the original
handler.

Second, an evaluation-replay scrub removes rejected tool calls from the
recorded action history before the benchmark evaluator replays actions
in a fresh environment. Without this scrub, a blocked call might be
replayed on an ungated evaluation environment and create a mismatch
unrelated to the actual gated run.

Both patches are no-ops when no gates are attached, preserving the
vanilla path.

\subsection{Conditions}\label{conditions}

The budget tier uses a gpt-4o-mini agent with a gpt-4.1 user simulator,
both at temperature 0, on the corrected airline task set: the current,
annotation-corrected \(\tau^2\)-bench airline tasks, which fix labeling
errors in the earlier release that otherwise cap measured performance,
loaded through the official \(\tau^2\)-bench registry \cite{barres2025tau2bench}.
We run all 50 tasks at 5 trials per task, for 250 trials per condition.

The frontier tier uses a gpt-5.2 agent at the harness default reasoning
level with the same gpt-4.1 user simulator. This configuration is not
intended to be leaderboard-comparable. Its in-harness vanilla baseline
is 61.2\%, and all frontier claims are measured against that baseline.

The headline gated condition is the four-gate suite described above. A
larger six-gate \texttt{optimal\_plus} suite is used only as a
budget-tier upper bound and is not the main configuration.

\subsection{Metrics and statistics}\label{metrics-and-statistics}

We report pass\(_1\) and the \(\tau\)-bench unbiased pass\(_k\)
estimator. pass\(_k\) measures the probability that all \(k\)
independent trials of a task succeed, averaged over tasks. It therefore
exposes whether success is reliable or occasional.

For significance, we use a paired bootstrap with the task as the
resampling unit. This preserves the paired structure between vanilla and
gated conditions. Reported intervals are 95\% percentile confidence
intervals from 20,000 resamples.

Gates add no model calls. Their runtime cost is limited to deterministic
reads and predicate evaluation.

\subsection{Leaderboard
non-comparability}\label{leaderboard-non-comparability}

The reported numbers should not be read as leaderboard submissions. The
gated scaffold modifies the agent environment, and the user simulator /
task-set configuration does not map cleanly onto a public board row. The
relevant comparison is therefore within-harness: vanilla versus the same
harness with gates.

\section{Results}\label{results}

\subsection{Aggregate lift and replication on the budget
model}\label{aggregate-lift-and-replication-on-the-budget-model}

On the full 50-task airline benchmark, the four-gate suite raises
gpt-4o-mini pass\(_1\) from 29.6\% to 42.0\%, a +12.4pp improvement
(Table~\ref{tab:budget-replication}).

\begin{table*}[t]
\small
\centering
\caption{Budget-model aggregate lift and disjoint-seed replication.}
\label{tab:budget-replication}
\begin{tabular}[]{@{}
  >{\raggedright\arraybackslash}p{(\textwidth - 4\tabcolsep) * \real{0.2727}}
  >{\raggedleft\arraybackslash}p{(\textwidth - 4\tabcolsep) * \real{0.3636}}
  >{\raggedleft\arraybackslash}p{(\textwidth - 4\tabcolsep) * \real{0.3636}}@{}}
\toprule\noalign{}
\begin{minipage}[b]{\linewidth}\raggedright
Condition
\end{minipage} & \begin{minipage}[b]{\linewidth}\raggedleft
Original n=5
\end{minipage} & \begin{minipage}[b]{\linewidth}\raggedleft
Replication n=15
\end{minipage} \\
\midrule\noalign{}
\bottomrule\noalign{}
Vanilla & 74/250 (29.6\%) & 231/750 (30.8\%) \\
Verified, four-gate suite & 105/250 (42.0\%) & 323/750 (43.1\%) \\
\(\Delta\) & +12.4pp, 95\% CI {[}+4.0, +21.2{]}, \(P=0.0012\) & +12.3pp,
95\% CI {[}+4.1, +21.3{]}, \(P=0.0008\) \\
\end{tabular}
\end{table*}

The replication is important. The original result uses five trials per
task, which is small enough to raise a seed-luck concern. The 15-seed
replication uses disjoint seeds and three times as many trials. The lift
reproduces to within 0.1pp, while the baseline reproduces to within
1.2pp. This supports the claim that the improvement is a stable property
of the intervention in this domain.

\subsection{The lift concentrates where gates
fire}\label{the-lift-concentrates-where-gates-fire}

Splitting the 50 tasks by whether any gate fires in the verified
condition isolates the mechanism (Table~\ref{tab:firing-budget}).

\begin{table*}[t]
\small
\centering
\caption{Budget-model stratification by whether any gate fires.}
\label{tab:firing-budget}
\begin{tabular}[]{@{}
  >{\raggedright\arraybackslash}p{(\textwidth - 8\tabcolsep) * \real{0.1579}}
  >{\raggedleft\arraybackslash}p{(\textwidth - 8\tabcolsep) * \real{0.2105}}
  >{\raggedleft\arraybackslash}p{(\textwidth - 8\tabcolsep) * \real{0.2105}}
  >{\raggedleft\arraybackslash}p{(\textwidth - 8\tabcolsep) * \real{0.2105}}
  >{\raggedleft\arraybackslash}p{(\textwidth - 8\tabcolsep) * \real{0.2105}}@{}}
\toprule\noalign{}
\begin{minipage}[b]{\linewidth}\raggedright
Stratum
\end{minipage} & \begin{minipage}[b]{\linewidth}\raggedleft
Tasks
\end{minipage} & \begin{minipage}[b]{\linewidth}\raggedleft
Vanilla
\end{minipage} & \begin{minipage}[b]{\linewidth}\raggedleft
Verified
\end{minipage} & \begin{minipage}[b]{\linewidth}\raggedleft
\(\Delta\)
\end{minipage} \\
\midrule\noalign{}
\bottomrule\noalign{}
Gate fires & 26 & 18/130 (13.8\%) & 43/130 (33.1\%) & +19.2pp, 95\% CI
{[}+6.9, +33.1{]}, \(P=0.0006\) \\
Gate never fires & 24 & 56/120 (46.7\%) & 62/120 (51.7\%) & +5.0pp, 95\%
CI {[}-5.0, +14.2{]}, \(P=0.18\) \\
\end{tabular}
\end{table*}

The firing stratum carries +25 of the +31 net successful trials. The
non-firing stratum moves by +5.0pp, but its interval includes zero. We
therefore do not claim that gates improve non-firing tasks. The
meaningful statement is that aggregate lift is concentrated where the
intervention is exercised.

\subsection{Suggestive frontier-model
evidence}\label{suggestive-frontier-model-evidence}

We also evaluate the same four-gate suite with a gpt-5.2 agent at the
harness default reasoning level. This arm is suggestive only: n=5 per
task, no replication, and not leaderboard-comparable.

The frontier agent's in-harness vanilla baseline is 61.2\% (153/250).
With gates, success rises to 71.6\% (179/250), a +10.4pp gain. A paired
task-level bootstrap gives a 95\% CI of {[}+0.4, +20.8{]} and
\(P=0.020\).

The gate fires on 43/250 verified frontier trials across 18 tasks, which
directly shows that the frontier model still attempts policy-violating
writes in this harness. However, because the frontier result is
unreplicated and the lower confidence bound is close to zero, it should
be read as supporting evidence rather than as the paper's statistical
anchor.

The same firing-concentration pattern appears at the frontier
(Table~\ref{tab:firing-frontier}):

\begin{table*}[t]
\small
\centering
\caption{Frontier-model stratification by whether any gate fires. Point estimates only.}
\label{tab:firing-frontier}
\begin{tabular}[]{@{}lrrrr@{}}
\toprule\noalign{}
Stratum & Tasks & Vanilla & Verified & \(\Delta\) \\
\midrule\noalign{}
\bottomrule\noalign{}
Gate fires & 18 & 30/90 (33.3\%) & 60/90 (66.7\%) & +33.3pp \\
Gate never fires & 32 & 123/160 (76.9\%) & 119/160 (74.4\%) & -2.5pp \\
\end{tabular}
\end{table*}

We do not attach a significance claim to this stratification. The point
estimate is included to show that the qualitative pattern recurs: lift
appears where gates fire, not where they do not.

\subsection{\texorpdfstring{Gates improve reliability under
pass\(_k\)}{Gates improve reliability under pass\_k}}\label{gates-improve-reliability-under-pass_k}

pass\(_1\) measures average success, but pass\(_k\) exposes consistency.
The budget model's vanilla performance collapses as \(k\) increases,
while gated configurations remain substantially higher
(Table~\ref{tab:passk}).

\begin{table*}[t]
\small
\centering
\caption{Reliability under pass$_k$ on the budget model.}
\label{tab:passk}
\begin{tabular}[]{@{}
  >{\raggedleft\arraybackslash}p{(\textwidth - 10\tabcolsep) * \real{0.1667}}
  >{\raggedleft\arraybackslash}p{(\textwidth - 10\tabcolsep) * \real{0.1667}}
  >{\raggedleft\arraybackslash}p{(\textwidth - 10\tabcolsep) * \real{0.1667}}
  >{\raggedleft\arraybackslash}p{(\textwidth - 10\tabcolsep) * \real{0.1667}}
  >{\raggedleft\arraybackslash}p{(\textwidth - 10\tabcolsep) * \real{0.1667}}
  >{\raggedleft\arraybackslash}p{(\textwidth - 10\tabcolsep) * \real{0.1667}}@{}}
\toprule\noalign{}
\begin{minipage}[b]{\linewidth}\raggedleft
k
\end{minipage} & \begin{minipage}[b]{\linewidth}\raggedleft
1
\end{minipage} & \begin{minipage}[b]{\linewidth}\raggedleft
2
\end{minipage} & \begin{minipage}[b]{\linewidth}\raggedleft
3
\end{minipage} & \begin{minipage}[b]{\linewidth}\raggedleft
4
\end{minipage} & \begin{minipage}[b]{\linewidth}\raggedleft
5
\end{minipage} \\
\midrule\noalign{}
\bottomrule\noalign{}
Vanilla & 29.6\% & 16.8\% & 11.6\% & 9.2\% & 8.0\% \\
Verified, four-gate suite & 42.0\% & 33.2\% & 29.4\% & 27.2\% &
26.0\% \\
Verified, six-gate upper bound & 46.0\% & 37.2\% & 34.0\% & 32.4\% &
32.0\% \\
\end{tabular}
\end{table*}

From \(k=1\) to \(k=5\), vanilla falls from 29.6\% to 8.0\%. The
four-gate suite falls from 42.0\% to 26.0\%. At \(k=5\), the four-gate
suite achieves more than three times the vanilla pass rate. This
supports the interpretation that gates remove a recurring failure mode
rather than merely adding random occasional wins.

\subsection{Gate precision and load-bearing
gates}\label{gate-precision-and-load-bearing-gates}

We audit gate rejections by comparing each blocked call to the
ground-truth trajectory. A true block is a rejected write that the
ground-truth trajectory also avoids. A false block is a rejected write
that the ground-truth trajectory performs.

\begin{table*}[t]
\small
\centering
\caption{Per-gate precision audit (pooled rejections) and minus-one removal effects over the five-gate candidate set; the four promoted gates (Table~\ref{tab:gates}) are the headline suite.}
\label{tab:gate-audit}
\begin{tabular}[]{@{}
  >{\raggedright\arraybackslash}p{(\textwidth - 10\tabcolsep) * \real{0.30}}
  >{\raggedleft\arraybackslash}p{(\textwidth - 10\tabcolsep) * \real{0.14}}
  >{\raggedleft\arraybackslash}p{(\textwidth - 10\tabcolsep) * \real{0.14}}
  >{\raggedleft\arraybackslash}p{(\textwidth - 10\tabcolsep) * \real{0.14}}
  >{\raggedleft\arraybackslash}p{(\textwidth - 10\tabcolsep) * \real{0.14}}
  >{\raggedleft\arraybackslash}p{(\textwidth - 10\tabcolsep) * \real{0.14}}@{}}
\toprule\noalign{}
\begin{minipage}[b]{\linewidth}\raggedright
Gate
\end{minipage} & \begin{minipage}[b]{\linewidth}\raggedleft
Fires
\end{minipage} & \begin{minipage}[b]{\linewidth}\raggedleft
True blocks
\end{minipage} & \begin{minipage}[b]{\linewidth}\raggedleft
False blocks
\end{minipage} & \begin{minipage}[b]{\linewidth}\raggedleft
Precision
\end{minipage} & \begin{minipage}[b]{\linewidth}\raggedleft
Removal \(\Delta\)
\end{minipage} \\
\midrule\noalign{}
\bottomrule\noalign{}
\texttt{cancellation\_eligibility} & 161 & 161 & 0 & 100\% & -2 \\
\texttt{must\_read\_before\_write} & 90 & 70 & 20 & 78\% & +3 \\
\texttt{baggage\_allowance} & 42 & 2 & 40 & 5\% & +3 \\
\texttt{basic\_economy} & 18 & 15 & 3 & 83\% & +6 \\
\texttt{passenger\_count} & 9 & 9 & 0 & 100\% & +4 \\
\end{tabular}
\end{table*}

Table~\ref{tab:gate-audit} audits the full candidate set of five gates,
pooled across runs; the removal \(\Delta\) for each gate is the
pass\(_1\) change when that gate is dropped from the full five-gate
bundle. The four gates in Table~\ref{tab:gates} are the promoted headline
suite, and \texttt{basic\_economy} is a fifth candidate that was
evaluated here but not promoted into it, the headline suite having been
fixed on the budget tier. The candidate set is not uniformly beneficial
gate-by-gate. The dominant success lift comes from
\texttt{cancellation\_eligibility}, which has 100\% precision in the
audit and is the only gate whose removal lowers pass\(_1\). Several
other gates block real violations but are not load-bearing on aggregate
task success in this distribution; one gate has poor precision and
should be treated as a candidate for revision before broader deployment.

The correct conclusion is not that every hand-written gate helps. The
correct conclusion is that high-precision deterministic gates over
policy-permissive writes can be strongly beneficial, and that gate
precision must itself be audited.

\subsection{Canonical deceptive task}\label{canonical-deceptive-task}

The cancellation gate is especially important on the canonical deceptive
task \#48, where the user provides false context to induce an
out-of-policy cancellation. On the budget and gpt-4.1 tiers, the
verified configuration solves this task 16/16 across model/settings
combinations, while non-gated conditions solve it at most 1/16. The gate
succeeds because cancellation eligibility is determined by database
state and policy, not by the user's asserted context.

This example illustrates the value of external verification. When a user
supplies misleading context, the model may accept it. A deterministic
gate reads the underlying state and blocks the forbidden write
regardless of the user's framing.

\subsection{Firing is necessary but not
sufficient}\label{firing-is-necessary-but-not-sufficient}

Gate firing does not guarantee recovery. A blocked violation can still
leave the agent unable to complete the task. The clearest budget example
is task \#39, where the gate fires repeatedly but the task remains 0/5.
The agent loops against the rejection rather than finding a compliant
plan.

This distinction matters. Gates provide a deterministic guarantee over a
proposed violating action, not over the full task. They remove a class
of silent state corruptions; recovery after rejection remains
model-dependent.

\subsection{Regressions}\label{regressions}

The gates also create regressions. At the frontier tier, three tasks
both fire gates and lose ground, indicating genuine gate-related harm.
These regressions are absorbed by larger gains in the firing stratum,
but they show that a suite tuned on one model may over-block another.

Other small losses occur on tasks with zero gate firings. Those should
not be attributed causally to gates; they are consistent with run-to-run
variation in a temperature-zero multi-agent harness.

The honest deployment lesson is that gates require per-policy and
per-model auditing. Determinism makes the gate decision reproducible,
but it does not automatically make every gate correct.

\section{Boundary and Scarcity}\label{boundary-and-scarcity}

\subsection{Negative control: self-enforcing retail
tools}\label{negative-control-self-enforcing-retail-tools}

In the \(\tau^2\)-bench retail domain, tools already enforce many
relevant preconditions. A forbidden call raises an error rather than
silently mutating state. A retail gate therefore duplicates checks that
exist in the tool layer.

On eight retail tasks, a retail gate suite moves performance from 40/64
(62.5\%) to 37/64 (57.8\%), a -4.7pp point estimate with a confidence
interval that includes zero. The negative point estimate is largely
explained by an encoding bug that blocked corrected retries the
underlying tool would have accepted.

This result supports the boundary claim. Where tools self-enforce, there
is no silent wrong-state class to recover. The right engineering
decision is to skip redundant gates or implement policy directly inside
the tool.

\subsection{Negative control: BFCL}\label{negative-control-bfcl}

On BFCL v4 \texttt{multi\_turn\_base} \cite{patil2025bfcl}, a
schema-existence gate produces zero firings across 200 entries.
Performance remains within run-to-run variance: 52.5\% vanilla versus
51.0\% gated. The dominant errors are wrong-sequence or instance-state
mismatches, not silent policy-violating writes. Bad calls generally
return structured errors rather than silently corrupting state.

BFCL therefore fails the same admission criterion: there is no
policy-permissive silent mutation for gates to intercept.

\subsection{Subset inflation}\label{subset-inflation}

The same run that yields +12.4pp on the full 50-task airline benchmark
yields +30.0pp on a curated eight-task subset enriched for the target
failure class. This shows two things at once.

First, the mechanism can be large when the task distribution is rich in
policy-permissive violations. Second, subset results can overstate
representative performance. The full-50 result is the appropriate
headline number; the curated subset is a useful upper bound under
failure-class enrichment.

The effect is also model-dependent. A subset selected for budget-model
failures need not be difficult for a frontier model. Firing share
follows the task distribution and model behavior, not merely the
benchmark label.

\subsection{Why a second positive domain is
difficult}\label{why-a-second-positive-domain-is-difficult}

A second positive domain would strengthen the paper substantially. We
searched for one and did not find an independent benchmark satisfying
all five admission axes. This is not a substitute for a second positive
result, but it is informative about the benchmark landscape.

WorkBench \cite{styles2024workbench} is the closest near-miss. It has
structured calls, permissive tools, final-state evaluation, and
state-decidable rules once a scored policy layer is authored. However,
its cooperative task distribution rarely induces violations. In a staged
probe over five state-reading gates, only one rule fires, and the
observed violation is an omission rather than a pressured decision to
break policy.

This result sharpens the fifth admission axis. A benchmark must not only
have permissive tools and state-decidable policies; it must also contain
tasks that induce the model to violate those policies. The scarcity of
such benchmarks limits the empirical scope of the present paper and
motivates future benchmark construction.

\section{Limitations}\label{limitations}

We collect the main limitations, several of which are discussed in context above.
(1)~\emph{One positive domain.} The headline result is a single domain (\(\tau^2\)-bench airline); the generality of the failure class remains a conjecture supported by one strong instance and two negative controls.
(2)~\emph{Suggestive frontier evidence.} The frontier arm is unreplicated (n=5, lower confidence bound near zero) and is not the paper's statistical anchor.
(3)~\emph{Uneven gate quality.} One gate carries most of the lift and one has poor precision, so gates require per-policy and per-model auditing.
(4)~\emph{No recovery guarantee.} A gate blocks a proposed violating write but does not ensure the agent recovers; post-rejection behavior is model-dependent.
(5)~\emph{Not leaderboard-comparable.} All reported numbers are within-harness comparisons under a specific user simulator and reasoning configuration.
(6)~\emph{State-decidability.} The method applies only when policy rules can be decided from current state and arguments; policies requiring ambiguity resolution, legal interpretation, or human judgment need richer mechanisms.
(7)~\emph{Untested baselines.} We do not test whether forcefully prompting the gated rules, or adding a reflection step, recovers the lift; for the deceptive cases at least (Section~\ref{canonical-deceptive-task}, where the user asserts a false timestamp) prompting cannot help, because the model is misled about state rather than ignorant of the rule, but the non-deceptive failures remain open.
(8)~\emph{Task overfitting.} Gates were written from the policy and evaluated on the same task set, and the replication is over seeds, not tasks; a held-out-task evaluation with frozen gates is the cleaner test, though the dominant gate's 100\% precision over 161 fires is partial evidence against task-fitting.
(9)~\emph{Rejection-message confound.} The structured reason a gate returns is itself a signal the agent re-plans on, so part of the recovery may come from that feedback rather than the block alone; the deterministic guarantee over the blocked write is unaffected, but isolating informative from generic rejection is future work.

\section{Discussion}\label{discussion}

The main lesson is that policy enforcement and task success are not
always in tension. In a policy-permissive environment, a forbidden write
can silently corrupt state and make the task fail. Blocking that write
can therefore improve final-state task success, not only safety.

The second lesson is that evaluation should distinguish loud errors from
silent wrong states. A benchmark that reports only pass/fail may hide
the operationally important difference between a tool call that fails
loudly and one that succeeds while violating policy. The latter is more
dangerous because the agent receives no corrective signal and the
transcript may appear successful.

The third lesson is that deterministic structure around LLM agents
remains valuable even as models improve. A stronger model may reduce the
frequency of violations, but a deterministic gate changes the system
property: the known forbidden transition is blocked whenever the gate
fires. This is a different kind of guarantee from probabilistic model
improvement.

At the same time, the evidence supports a bounded claim: gates help only
where the policy is state-decidable, the tool is permissive, and the task
distribution exercises the violation, and are redundant or insufficient
otherwise.

\section{Conclusion}\label{conclusion}

We identify silent policy violations on policy-permissive tools as a
trust-relevant failure mode in tool-using LLM agents. In the
\(\tau^2\)-bench airline domain, many failures are silent wrong states
with no tool error, and the aggregate failure rate is reproducible
across disjoint seeds rather than incidental. A lightweight suite of deterministic,
read-only pre-execution gates recovers a substantial fraction of these
failures, lifting gpt-4o-mini success from 29.6\% to 42.0\% and
reproducing the lift on a disjoint 15-seed replication set.

The result is best understood as a bounded evaluation and reliability
contribution. Gates do not solve agent safety generally, and they do not
guarantee task success. They do provide a deterministic block on a known
class of policy-violating writes at the moment those writes are
proposed. Where tools are policy-permissive and policies are
state-decidable, that block can turn a silent unrecoverable corruption
into an explicit rejection the agent can recover from.

This suggests a concrete direction for agent evaluation: benchmarks and
deployment harnesses should expose whether failures are loud or silent,
whether policies live in prompts or tools, and whether deterministic
verification at the action boundary can prevent state corruption before
it occurs.

\balance
\bibliographystyle{ACM-Reference-Format}
\bibliography{refs}

\end{document}